\documentclass{article} 
\usepackage{iclr2026_conference,times}
\usepackage{subfig}

\usepackage{amsmath,amsfonts,bm}

\def\eqref#1{equation~\ref{#1}}

\def\1{\bm{1}}

\DeclareMathAlphabet{\mathsfit}{\encodingdefault}{\sfdefault}{m}{sl}
\SetMathAlphabet{\mathsfit}{bold}{\encodingdefault}{\sfdefault}{bx}{n}

\usepackage{hyperref}
\usepackage{url}
\usepackage[nolist,nohyperlinks]{acronym}
\usepackage{graphicx}

\definecolor{peach}{HTML}{F26035}

\begin{acronym}
    \acro{AL}[AL]{active learning}
    \acro{PRE}[PRE]{physics residual error}
    \acro{PDE}[PDE]{partial differential equation}
    \acro{CPU}[CPU]{central processing unit}
    \acro{GPU}[GPU]{graphical processing unit}
    \acro{FNO}[FNO]{Fourier neural operator}
    \acro{LCMD}[LCMD]{largest cluster maximum distance}
    \acro{IC}[IC]{initial condition}
    \acro{PINN}[PINN]{physics-informed neural network}
    \acro{SBAL}[SBAL]{stochastic batch active learning}
    \acro{FFT}[FFT]{fast Fourier transform}
    \acro{RMSE}[RMSE]{root mean square error}  
    \acro{UQ}[UQ]{uncertainty quantification}
\end{acronym}

\title{Data-Efficient Neural Operator Training via Physics-Based Active Learning}

\author{%
  Alicja~Polanska\\
  University College London\\
  United Kingdom Atomic Energy Authority\\
  \texttt{alicja.polanska.22@ucl.ac.uk} \\
  \And
  Lorenzo~Zanisi \\
  United Kingdom Atomic Energy Authority \\
  \texttt{lorenzo.zanisi@ukaea.uk} \\
  \AND
  Vignesh~Gopakumar \\
  United Kingdom Atomic Energy Authority \\
  University College London\\
  \texttt{vignesh.gopakumar@ukaea.uk} \\
  \And
  Stanislas Pamela \\
  United Kingdom Atomic Energy Authority \\
  \texttt{stanislas.pamela@ukaea.uk} \\
}

\iclrfinalcopy 
\begin{document}

\maketitle

\begin{abstract}
Solving partial differential equations with neural operators significantly reduces computational costs but remains bottlenecked by high training data requirements. Active learning offers a natural framework to mitigate this by selectively acquiring the most informative samples in an iterative manner. We introduce physics-based acquisition -- a novel physics-informed active learning algorithm that leverages the partial differential equation residual to guide data selection. We validate the method by presenting numerical experiments for the 1D Burgers equation and the 2D compressible Navier-Stokes equations. We show that, in our experiments, physics-based acquisition consistently outperforms random acquisition and matches the state of the art in data efficiency. At the same time, it has the unique advantage of injecting a physics inductive bias into the training process, ensuring that simulation cost is spent where the model’s physical understanding is weakest.
\end{abstract}

\section{Introduction}
In the physical sciences, \acp{PDE} are the language used to describe complex dynamical systems. From modelling plasma dynamics in fusion reactors to simulating galaxy formation, we rely on numerical solvers to predict the evolution of complex and highly parameterised \ac{PDE} systems. However, numerical solvers can be extremely computationally expensive, which prevents their use in highly iterative control and optimisation workflows. For example, high-fidelity plasma simulations can take several weeks if not months on modern supercomputers \citep{Smith_2020}, galaxy formation simulations require millions of CPU hours \citep{10.1093/mnras/stx3112}. To alleviate this computational burden, many machine learning-based methods for approximating  \ac{PDE} solution operators have been proposed, most notably neural operator surrogate models \citep{li2020no, 2023KovachkiNOs, Lu_2021,Takamoto2022pdebench}.

Neural operators require large training sets to obtain surrogate models that perform well for a desired range of \ac{PDE} parameters, initial and boundary conditions. As the training data comes from the very simulators the surrogate models are supposed to replace, we are posed with the proverbial `chicken-and-egg' problem in surrogate modelling \citep{brandstetter2022liepointsymmetrydata}: for very expensive simulators, it is essential to reduce the size of the training sets to make surrogate modelling feasible in the first place.
\Ac{AL} offers a natural solution to this problem. As a paradigm designed to minimise labelling costs, \ac{AL} iteratively selects the most informative unlabelled data points to add to the training set \citep{mackay_1992, ren2021surveydeepactivelearning, li2024surveydeepactivelearning}. This approach is especially useful in settings where obtaining data labels carries a high cost, making it ideal for improving the data efficiency of neural operators. \ac{AL} for \acp{PDE} has been utilised in previous studies for simplified tasks such as the prediction of univariate outputs, or fixed time point solutions \citep{pestourie2020active, pickering2022discovering, wu2023disentangled, bajracharya2024feasibility, wu2023deep}. Recently, \ac{AL} has also been explored for the task of full spatio-temporal trajectory prediction with neural operators \citep{al4pde-benchmark-musekamp:2024, kimactive}. 

Most of these works attempt to adapt standard \ac{AL} tools to \ac{PDE} learning by, e.g., prioritising acquisition of simulations where an ensemble of surrogates produces the highest variance \citep{pestourie2020active, pickering2022discovering}, information-theoretic arguments \citep{wu2023deep, wu2023disentangled}, variance reduction methods \citep{kimactive} or clustering \citep{al4pde-benchmark-musekamp:2024}. Instead, we notice that the residual of the \ac{PDE} provides a viable metric over the solution accuracy of the neural surrogate. Solutions with high residual values often tend to deviate further from the physical solution. 
 \ac{PDE} residuals are utilised widely for machine learning frameworks in the context of \acp{PINN} \citep{raissi2019physicsPRE} and have been the main workhorse of \ac{PDE} analysis and optimisation within numerical and computational physics \citep{numerical_PDEs}. In recent work, the \ac{PRE} was introduced as an uncertainty measure for neural operators  \citep{gopakumar2025calibrated}. The \ac{PRE} provides a natural and physically-grounded acquisition function for \ac{AL}: we can improve the surrogate by querying simulations where its current predictions are the most unphysical. Concurrently, \ac{PDE} residuals have been utilised for single-shot coreset selection for neural operator training \citep{satheesh2025picore}. However, such static approaches cannot adapt as the surrogate model improves during training.

Our contributions are as follows: We introduce a novel physics-based acquisition strategy that utilises the \ac{PDE} residual as a principled physics-informed measure of model epistemic uncertainty; We integrate this strategy into the \textsc{al4pde} framework \citep{al4pde-benchmark-musekamp:2024}, providing a robust benchmark against established methods; We demonstrate that our physics-based approach achieves competitive data efficiency with state-of-the-art methods, while injecting a physics inductive bias. 

\section{Methodology}
\subsection{Active Learning for Neural Operators}
Neural operators learn mappings between infinite-dimensional function spaces. For a given family of \acp{PDE}, the model learns a discretisation-independent solution operator. Although many neural operator architectures have been devised in recent years \citep{cao2023lnolaplaceneuraloperator,alkin2025universalphysicstransformersframework,serrano2024aromapreservingspatialstructure}, in this short proof-of-concept work, we focus on 
\acp{FNO} \citep{li_kovachki_azizzadenesheli_liu_bhattacharya_stuart_anandkumar_2020} due to their efficiency and cost-accuracy trade-off \citep{dehoop2022costaccuracytradeoffoperatorlearning}. The \ac{FNO} architecture is based on spectral convolutions, where the input to each layer is first converted to the frequency domain via a \ac{FFT}, then multiplied by a weight matrix in Fourier space, and finally transformed back to the original domain using an inverse \ac{FFT} \citep{li_kovachki_azizzadenesheli_liu_bhattacharya_stuart_anandkumar_2020}. We condition the model on the \ac{PDE} parameters by appending the raw parameter values as additional constant channels to the
model input \citep{takamoto2023learningneuralpdesolvers}, following the \textsc{al4pde} implementation \citep{al4pde-benchmark-musekamp:2024}. 

\Ac{AL} algorithms utilise acquisition functions to sample new training data iteratively from a data pool and have demonstrated data-efficiency gains across classification \citep{ren2021surveydeepactivelearning} and regression \citep{holzmuller2023framework} compared to the random sampling baseline. \textsc{al4pde} is a recently introduced extensible framework for the development and evaluation of \ac{AL} algorithms for neural \ac{PDE} solvers \citep{al4pde-benchmark-musekamp:2024}. It is implemented as an open-source Python package\footnote{\url{https://github.com/dmusekamp/al4pde}}, providing solvers for a variety of parametric \acp{PDE}, neural surrogate models and state-of-the-art \ac{AL} algorithms borrowed from the literature on \ac{AL} for standard regression problems \citep{holzmuller2023framework}. Specifically, acquisition functions pick sets of \acp{IC} and \ac{PDE} parameters by exploiting surrogate solutions obtained autoregressively from trained surrogate models. Two notable acquisition schemes that achieve state-of-the-art results are \ac{LCMD} \citep{holzmuller2023framework} and \ac{SBAL} \citep{kirsch2023SBAL}, clustering-based and uncertainty-based AL methods, respectively.

\subsection{Physics Residual Error}
We extend the \textsc{al4pde} framework with a new acquisition function based on the physics residual error \citep{gopakumar2025calibrated}. We can define a \ac{PDE} governing the dynamics of $n$ field variables $u\in\mathbb{R}^n$ through the equations:
\begin{align}
    D = D_t(u) + D_X(u; \delta) &= 0, \quad X\in\Omega,\; t\in[0,T], \label{eqn:pde} 
\end{align}
where $X$ is the spatial domain bounded by $\Omega$, $[0,T]$ the temporal domain, $D_X$ and $D_t$ the composite operators of the associated spatial and temporal derivatives and $\delta$ indicates the parameters that appear in the PDE. The solution is further determined by the boundary condition $g$ and initial condition $a$, parameterised by $\lambda$:
\begin{align}
    u(X,t) &= g, \quad X\in\partial\Omega, \label{eqn:bc}\\
    u(X,0) &= a(\lambda, X) \label{eqn:ic}.
\end{align}
The \ac{PDE} residual is the evaluation of the composite differential operator $D$ over an approximate solution $\hat{u}$ \citep{morton2005numerical}:
\begin{equation}
\label{eq:D_PRE}
 D(\hat{u})-R=0.
\end{equation}
For an exact solution, $R=0$ everywhere in the domain. For approximate solutions from a numerical solver or surrogate model, it gives a quantitative measure of how physical the solution is. The \ac{PRE} is defined as the residual $R$ estimated over the discretised solution to the \ac{PDE}. PRE has been used as a measure of convergence and stability in PDE analysis across numerical methods \citep{numerical_PDEs}. Recently, it has been adopted as an uncertainty measure for neural operators deployed as surrogate models for a well-defined family of PDEs \citep{gopakumar2025calibrated}. 

To estimate the \ac{PRE}, we followed the method deployed in \citep{gopakumar2025calibrated}\footnote{\url{https://github.com/gitvicky/CP-PRE}}. Finite difference stencils were deployed as convolutional kernels, rendering the \ac{PRE} estimation as an additive convolutional operation. This method allows for gradient estimation without having access to the computational graph or model parameters and can be accelerated on GPUs.

\subsection{Physics-Based Acquisition}
\label{sec:PRE}
Our physics-based acquisition strategy leverages the \ac{PRE} as a direct measure of the physical accuracy of the surrogate model. For each candidate pair of initial conditions and \ac{PDE} parameters in the pool $(\delta, \lambda)$, we first rolled out the surrogate model to generate a trajectory. We then calculated an acquisition score $s(\delta, \lambda)$, by taking the mean absolute \ac{PRE} of the trajectory:
\begin{equation}
\label{eq:acquisition}
    s(\delta, \lambda) = \frac{1}{N_s N_t} \sum_{i=1}^{N_s} \sum_{t=1}^{N_t} |\text{PRE}(\mathbf{x}_i, t)|,
\end{equation}
where $\mathbf{x}_i$ represents the spatial grid points, $t$ the temporal steps, and $N_s$ and $N_t$ the total number of spatial and temporal points, respectively.

We employed two popular uncertainty-based acquisition strategies: \textsl{top k} and stochastic batch active learning (\textsc{SBAL}) \citep{kirsch2023SBAL}. For \textsl{top k} acquisition, we ranked the members of the pool by their score, and added the $k$ highest ranked to the training set. For \textsc{SBAL}, pool data were acquired into the training set in a stochastic manner, by perturbing the scores in \eqref{eq:acquisition} with a power-law noise distribution. This method has been shown to help diversify the acquisition without the need for bespoke batch \ac{AL} algorithms \citep{kirsch2023SBAL}.

The PDE residuals cannot be directly compared across different parameter choices because changes in coefficients alter the natural scaling and conditioning of the equations, making the same residual magnitude represent different levels of error. In order to account for varying dynamical regimes, we normalised each candidate trajectory’s acquisition score by that of the ground-truth trajectory corresponding to the closest training-set member in \ac{PDE} parameter space, measured by Euclidean distance \citep{Ferziger2020Computational}.

\section{Results}
\begin{figure}
    \centering
    \subfloat[Burgers \label{fig:RMSE_Burgers}]{\includegraphics[width=0.45\linewidth]{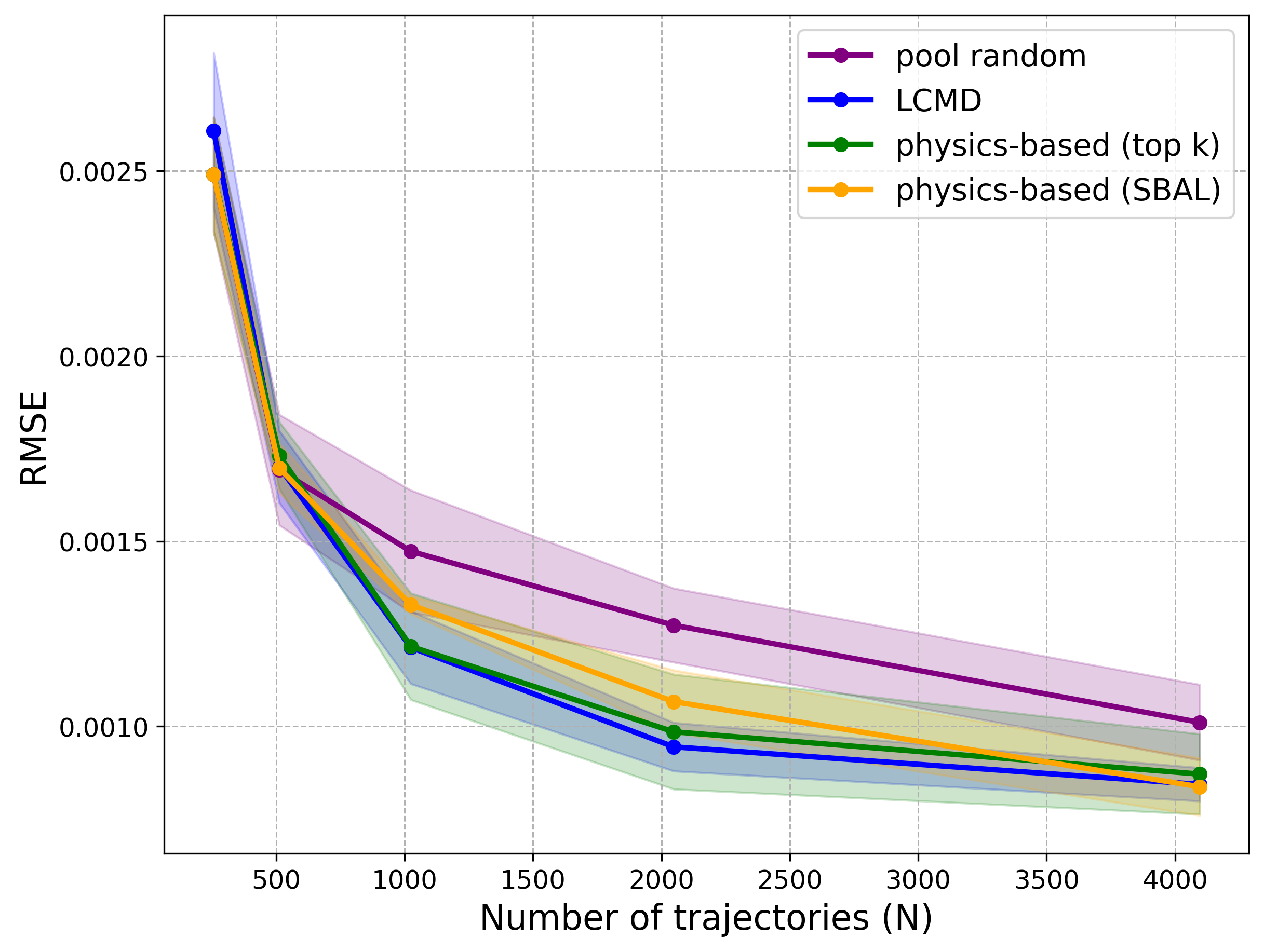}}
    \qquad
    \subfloat[2D Compressible Navier Stokes \label{fig:RMSE_NS}]{\includegraphics[width=0.45\linewidth]{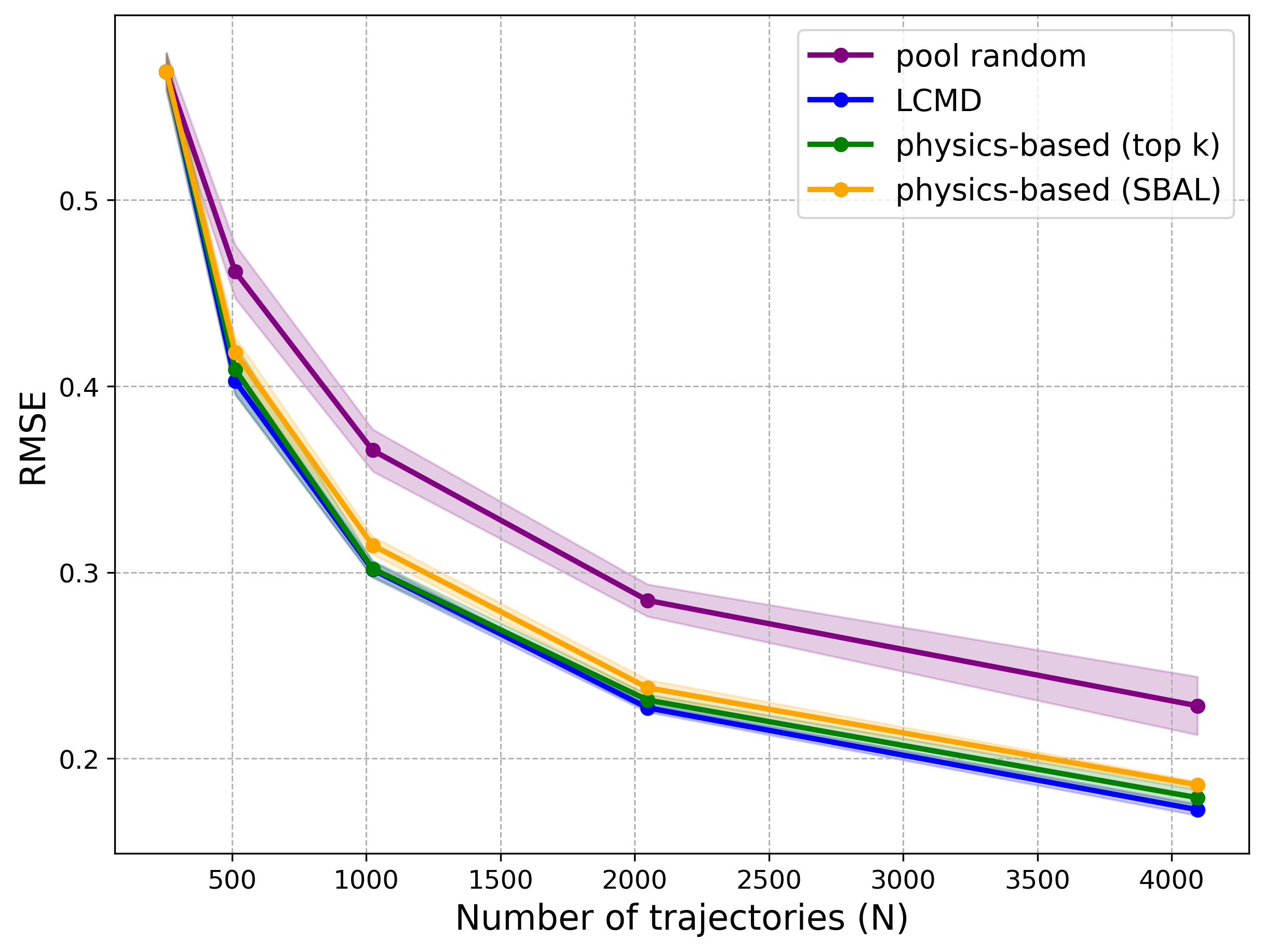}}
    \caption{\ac{RMSE} as a function of the number of training trajectories (N) for (a) the 1D Burgers equation and (b) 2D compressible Navier-Stokes equation. Solid lines show mean \ac{RMSE} averaged over multiple seeds, with shaded regions indicating $95\%$ confidence intervals. Our method significantly outperforms random acquisition and achieves similar data efficiency to \ac{LCMD}.}
    \label{fig:RMSE_v_N}
\end{figure}

To validate its efficiency, we implemented our novel physics-based acquisition algorithm in the \textsc{al4pde} benchmark \citep{al4pde-benchmark-musekamp:2024}. We made use of their \ac{AL} experiment set-up, focusing on the Burgers and 2D compressible Navier-Stokes systems. For details of the architecture and experimental setup, as well as the relevant equations, we refer the reader to \citep{al4pde-benchmark-musekamp:2024}. All experiments were executed on a single NVIDIA H100 SXM 80GB GPU. For both examples, we evaluated the model performance measured by \ac{RMSE} on the test set as a function of the number of trajectories in the training set $N$ (Figure~\ref{fig:RMSE_v_N}). Our physics-based acquisition strategy was benchmarked against random sampling (pool random) and \ac{LCMD}. We repeat the experiment for five different random seeds and plot the average over these runs. 

Figure~\ref{fig:RMSE_Burgers} shows the comparison between random, \ac{LCMD} and our physics-based acquisitions for the 1D Burgers equation for different acquisition schemes. The pool range for the kinematic viscosity was set to $\nu \in [0.1, 1)$. Figure \ref{fig:RMSE_NS} shows an analogous plot for the 2D compressible Navier-Stokes equation for different acquisition schemes for the shear and bulk viscosities $\eta, \zeta \in [10^{-2}, 10^{-1})$. It can be seen that our acquisition strategy consistently outperforms random sampling, achieving similar performance for a significantly lower number of training trajectories. The performance is on par with \ac{LCMD}, which was the best performing method in the benchmark of \citep{al4pde-benchmark-musekamp:2024}.

Our results establish that physics-based acquisition is a competitive alternative to purely data-driven strategies on parameter ranges that correspond to moderately turbulent (for compressible Navier-Stokes) and diffusion-dominated (for Burgers) cases. We identify two primary factors that may influence performance for wide parameter ranges: limitations in the \ac{FNO}'s conditioning scheme and availability of sufficient data for accurate \ac{PRE} normalisation. In future work we plan to explore more parameter-aware normalisation schemes, for instance based on theoretical scaling laws for gradients. While geometric methods like \ac{LCMD} provide broad coverage of the \ac{PDE} parameter space, our acquisition offers a unique physics-informed refinement of the model.

\section{Conclusions}

In this work, we introduced physics-based acquisition, a novel physics-informed \ac{AL} strategy for generating training datasets from costly \ac{PDE} simulators efficiently, leveraging the \ac{PRE} as an uncertainty measure. By preferentially acquiring training data for which the surrogate model produces the most unphysical solutions, our method actively guides the model towards adhering to the governing \ac{PDE}. Our experiments on the Burgers and 2D compressible Navier-Stokes equations demonstrate that our acquisition strategy significantly improves data efficiency compared to random sampling and achieves performance competitive with established, purely data-driven \ac{AL} methods like \ac{LCMD}. This capability is crucial for mitigating the high computational cost associated with generating large training datasets for neural operators. Our findings demonstrate the significant potential of injecting a physics inductive bias to steer data acquisition in complex, compute-bound physics domains.

Our current normalisation strategy utilises local ground-truth information to calibrate the \ac{PRE} across \ac{PDE} parameters. Our results demonstrate, that using the \ac{PRE} provides a strong signal for physics inconsistency, but robust normalisation is needed to ensure this signal is directly comparable across the parameter space for broad parameter range acquisition. While this makes acquisition for solutions across distinct dynamical regimes a challenge, it also makes the method particularly well-suited for applications where dynamics vary continuously with respect to \ac{PDE} parameters or where selecting initial conditions is the primary goal. In future work, we plan to further investigate and refine the normalisation scheme, as well as apply this methodology to more complex, real-world examples, for instance simulations of plasma dynamics. 


\bibliography{refs}
\bibliographystyle{iclr2026_conference}


\end{document}